# Variational Quantum Classifiers for Natural-Language Text

Daniel T. Chang (张遵)

*IBM (Retired)* dtchang43@gmail.com

**Abstract:** As part of the recent research effort on quantum natural language processing (QNLP), variational quantum sentence classifiers (VQSCs) have been implemented and supported in lambeq / DisCoPy, based on the DisCoCat model of sentence meaning. We discuss in some detail VQSCs, including category theory, DisCoCat for modeling sentence as string diagram, and DisCoPy for encoding string diagram as parameterized quantum circuit. Many NLP tasks, however, require the handling of text consisting of multiple sentences, which is not supported in lambeq / DisCoPy. A good example is sentiment classification of customer feedback or product review. We discuss three potential approaches to variational quantum text classifiers (VQTCs), in line with VQSCs. The first is a weighted bag-of-sentences approach which treats text as a group of independent sentences with task-specific sentence weighting. The second is a coreference resolution approach which treats text as a consolidation of its member sentences with coreferences among them resolved. Both approaches are based on the DisCoCat model and should be implementable in lambeq / DisCoCat. The third approach, on the other hand, is based on the DisCoCirc model which considers both ordering of sentences and interaction of words in composing text meaning from word and sentence meanings. DisCoCirc makes fundamental modification of DisCoCat since a sentence in DisCoCirc updates meanings of words, whereas all meanings are static in DisCoCat. It is not clear if DisCoCirc can be implemented in lambeq / DisCoCat without breaking DisCoCat.

## 1 Introduction

*Quantum Natural Language Processing (QNLP)* is a new area of research and applications [10-12]. It exploits and applies *quantum computing* to critical aspects of NLP, involving different NLP tasks. A major approach to QNLP for *near-term quantum computers* [10-11] aims at representing *sentence meanings* as vectors encoded into *parameterized quantum circuits (PQCs)* [1]. To achieve this, the distributional meaning of *words* is extended by the compositional meaning of *sentences,* based on the *DisCoCat (Distributional Compositional Categorical)* model [3] in which the vectors representing *word meanings* are composed to form sentence meaning through the *syntactic structure* of the sentence.

As part of the QNLP effort based on DisCoCat, *variational quantum sentence classifiers (VQSCs)* [13-14] have been implemented and supported in lambeq / DisCoPy. VQSCs use a *hybrid quantum-classical approach* in that the *parameters* of PQCs are classical objects and optimized using *classical optimization techniques*. Note that training VQSCs amounts to learning the sentence representations / encodings in a task-specific way, i.e., it is a form of *task-specific quantum metric learning* [2]. This means that the *optimized PQCs* can be used for other NLP tasks downstream, e.g. *sentence generation task* [14].

In this paper we discuss in some detail VQSCs, including category theory [4], DisCoCat for modeling sentence as string diagram, and DisCoPy [7, 9] for encoding string diagram as PQC. *DisCoCat* is a linguistic application of *category theory*, which is the mathematical study of (abstract) algebras of *functions*, consisting of *objects* A,B,C, . . . and *arrows* f : A → B, g : B → C, . . . DisCoCat comes with a pictorial representation, allowing any sentence to be represented by a *string diagram*. Such a diagram consists of *boxes* representing words, and *wires* connecting these boxes representing syntactic relations between words. In DisCoPy, *Diagram* (for string diagram) is a subclass of Arrow (for arrow), and *Circuit* (for quantum circuit) is a subclass of Diagram, thus facilitating the encoding of string diagram as PQC. *lambeq* [6] provides end-to-end support for VQSCs: from sentences to string diagrams to PQCs to training and evaluation of PQCs.

Many NLP tasks, however, require the handling of *text consisting of multiple sentences*, which is not supported in lambeq/ DisCoPy. A good example is sentiment classification of customer feedback or product review.

We discuss three potential approaches to *variational quantum text classifiers (VQTCs)*, in line with VQSCs. The first is a *weighted bag-of-sentences* approach which treats text as a group of independent sentences with task-specific sentence weighting. The second is a *coreference resolution* approach which treats text as a consolidation of its member sentences with coreferences among them resolved. Both approaches are based on the *DisCoCat* model.

The third approach, on the other hand, is based on the *DisCoCirc* model which considers both ordering of sentences and interaction of words in composing *text meaning* from word and sentence meanings. DisCoCirc makes fundamental modification of DisCoCat since a sentence in DisCoCirc *updates* meanings of words, whereas all meanings are *static* in DisCoCat.

## 2 DisCoCat for Modeling Sentence as String Diagram

The *Distributional Compositional Categorical (DisCoCat)* model of *sentence meaning* [3] is a mathematical framework that allows for the meaning of a *sentence* to be described as a composition of the meanings of its constituent *words*, based on the *grammatical relationships* between these words. This is in contrast to language models, such as "bags of words", which ignore the grammatical structure. DisCoCat is a linguistic application of *category theory* [4], which is discussed in Appendix A. Category Theory

DisCoCat comes with a pictorial representation, allowing any sentence to be represented by a *string diagram*. Such a diagram consists of *boxes* representing *words*, and *wires* connecting these boxes according to the formalism of *pregroup* *grammars* [5]. Every wire in the diagram is annotated either by some *atomic type p*, a *left adjoint $p^l$*, or a *right adjoint $p^r$*. In pregroup grammars, each *word* is a morphism with type $I \to T$ where I is the monoidal unit and T is a (rigid) pregroup type. Here are some examples for pregroup type assignments:

- a *noun* is given the *base type n*
- an *adjective* consumes a noun on the noun's left to return another noun, so it is given the *type $n \cdot n^l$*
- a *transitive verb* consumes a noun on its left and another noun on its right to give a *sentence*, so is given the *type $n^r \cdot s \cdot n^l$*

The following example is a *string diagram* for the sentence "John walks in the park":

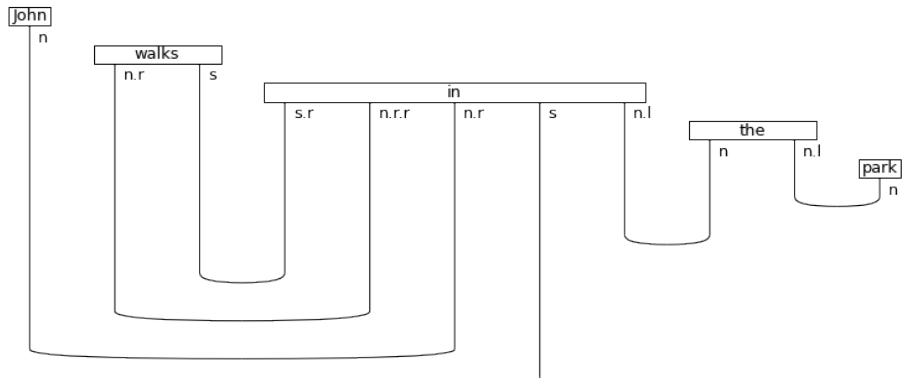

DisCoCat diagrams are based on the rigorous formalism of *monoidal categories* [3], which means they are equipped with a *diagrammatic calculus*. This calculus can be used to *rewrite* complicated string diagrams into simpler ones that still encode the meaning of the original sentence. This is necessary because syntactic derivations in pregroup form can become extremely complicated, which may lead to excessive use of hardware resources and prohibitively long training times. *lambeq* [6], a Python Library for QNLP with dependency on *DisCoPy* [7] and *Pytket* [8], comes with a number of standard *rewrite rules* covering auxiliary verbs, connectors, coordinators, adverbs, determiners, relative pronouns, and prepositional phrases, for example:

- determiner: removes determiners (such as "the") by replacing them with caps
- prepositional_phrase: simplifies prepositions by passing through the noun wire using a cap

The following is the *simplified string diagram* for the sentence "John walks in the park" after a series of rewrites:

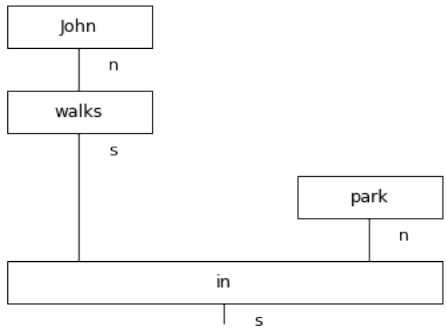

*lambeq* comes with a default parser *BobCatParser* with *sentence2diagram()* and *sentences2diagrams()* methods, which return string diagram(s) as object(s) of the DisCoPy *Diagram* class. This is the core class for *NLP* and *quantum computing* and is discussed in some detail in the next section. lambeq also comes with a *Rewriter*.

## 3 DisCoPy for Encoding String Diagram as Parameterized Quantum Circuit

*DisCoPy (Distributional Compositional Python)* [7, 9], the backend of *lambeq*, is a Python toolbox for computing with *string diagrams* and *functors*. The string diagrams are used to encode various kinds of *quantum processes*, with functors for compilation and evaluation on quantum hardware. Their implementation is given by *Diagram*, a class with the following interface [9] (see also A.1 DisCoPy Implementation of Category Theory):

• Diagram.*dom* and Diagram.*cod* are attributes of type *Ty*, a class which stores the list of labels for the input and output wires, also called the *domain* and *codomain*,

• Diagram.*id*, shortened to *Id*, is a static method which takes an x:Ty as input and returns the identity diagram with dom = cod = x,

• Diagram.*then* and Diagram.*tensor*, shortened to >> and @, are methods which take two diagrams and return their *composition* in sequence and in parallel respectively,

• Diagram.*dagger*, shortened to *[::-1]*, returns the *vertical reflection* of a diagram,

• Diagram.*draw* is a method which displays the diagram on screen or outputs code to be included in documents.
.

The Diagram class is general enough to encode any *string diagram*, and in particular any *quantum circuit*. Its subclass *Circuit* comes with methods *from_tk()* and *to_tk()* for back and forth conversion into the *Pytket Circuit* [8]. The Pytket Circuit format provides additional functionality and allows interoperability. For example, obtaining *Qiskit Circuit* is trivial by calling the *tk_to_qiskit()* method provided by the *Pytket-Qiskit extension*.

The following is a a procedure [6] for transforming any *string diagram* into a *parameterized quantum circuit (PQC)* [1] that can be run on *IBM Quantum* hardware or simulators:

1. An *ansatz* is used to transform a *simplified string diagram* to a PQC. This ansatz is a mapping that assigns a number of *qubits* to each wire type in the string diagram, as well as a set of *quantum gates* to each *box* in the diagram. DisCoPy comes with the class *IQPAnsatz* [10], which turns the string diagram into a standard *IQP (Insantanoues Quantum Polynomial) circuit*. The IQP circuit constitutes of one or more layers, each layer consisting of a row of *Hadamard gates*, followed by a ladder of *parameterized controlled-Z rotations*. At the end, a final row of *Hadamard gates* is applied. Note that the choice of ansatz determines the *parametrization* of the concrete family of models.
2. The quantum compiler t|ket⟩ is used to translate the PQC into *machine-specific instructions*, which can be executed on IBM quantum hardware or simulators. This is done by creating a lambeq *TketModel*, passing in the PQC and specifying a backend (e.g., the AerBackend provided by the *Pytket-Qiskit extension*).

The *DisCoPy PQC* corresponding to the *simplified string diagram* for the sentence "John walks in the park" is given below, created by assigning 1 qubit to the noun type and 1 qubit to the sentence type as well as setting the number of IQP layers to 2. The PQC has 11 (classical) *parameters*: 3 each associated with "John", "walks" and "park", and 2 associated with "in".

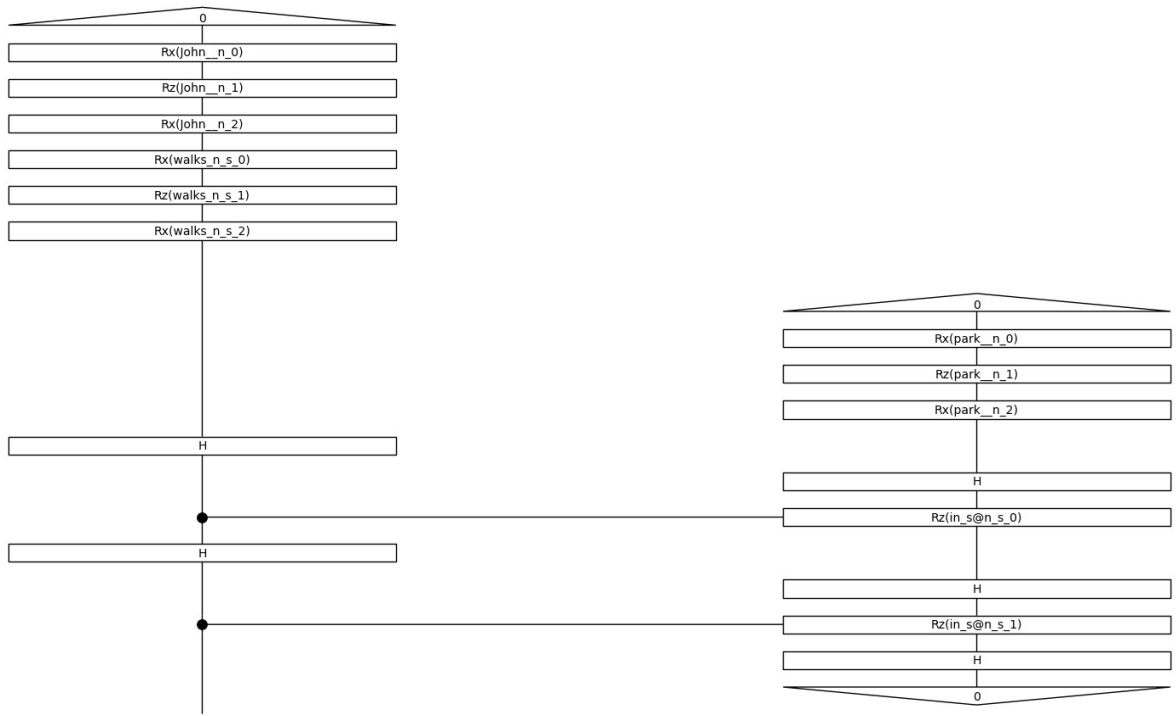

## 4 Variational Quantum Sentence Classifiers

The PQCs created in the last section can be used for various *quantum NLP* tasks, in a *variational, hybrid quantum-classical approach* because the *circuit parameters* are classical objects and their values are determined variationally using *classical optimization* techniques. For the current study, we focus on *quantum sentence classification* tasks [13-14]: Given a dataset Γ of *sentences*, each of which belongs to one of *k possible topics*, train a *classifier* that can correctly determine the topic of unseen sentences which are also about one of the k possible topics.

The algorithms involved for the *variational quantum sentence classifier (VQSC)* are as follows [14]:

1. Each *sentence S ∈ Γ* is converted to a *PQC $C_S$* using the techniques discussed in the previous section. Note that some parameters may be *shared* between PQCs corresponding to different sentences. This occurs when the *same words* appear in multiple sentences.
    - We set $q_n = 1$, and $q_s = \lceil log\ k \rceil$, where $q_n$ and $q_s$ are the number of qubits associated to the noun and sentence types respectively.

- *Measuring* such a circuit yields one of *k possible outcomes*, each of which is associated with one of the topics in the corpus.

2. For each sentence S and each *topic* $i \in \{0, 1, ..., k - 1\}$ we define a *binary predicate* $L(i, S) \in \{0, 1\}$ and set $L(i, S) = 1$ if and only if sentence S has topic i. Moreover, we define $P(i, C_S)$ as the probability of observing outcome i when measuring the final state of a PQC $C_S$.
    - Let $\Omega$ denote the full set of *parameters* used in all the PQCs combined. Our goal is to find the *optimal* $\Omega$ which *maximizes* $P(i, C_S)$ whenever $L(i, S) = 1$.
    - This can be solved using *classical machine learning techniques*, by minimizing the *categorical cross-entropy loss function* below, using the *SPSA (Simultaneous perturbation stochastic approximation) algorithm*:

    $$C(\Omega) = \Sigma_{S \in \Gamma} L(i, S) \bullet \log P(i, C_S)$$

3. Given an *unseen sentence* $S' \notin \Gamma$ we can predict its topic as follows: Use the *optimal parameters* $\Omega$ to create the *PQC* $C_{S'}$. Measure the final state of $CS'$ obtaining an *outcome* $i \in \{0, 1, ..., k - 1\}$. Output the *topic* associated with outcome i.

Note that training the model amounts to learning the *sentence representations* in a task-specific way, i.e., it is a form of *task-specific quantum metric learning* [2].

## 4.1 lambeq Support

*lambeq* [6] provides convenient high-level support for *supervised QNLP tasks*, including *quantum sentence classification* tasks. Specifically, it contains the following high-level abstract classes and several concrete implementations for them:

- *Model*: A model bundles the basic attributes and methods used for training, given a specific backend. It stores the *symbols* (for model parameters) and the corresponding *weights*, and implements the *forward pass* of the model. Concrete implementations include the *TketModel* class mentioned in the last section.
- *Optimizer*: An optimizer calculates the *gradient* of a given *loss function* with respect to the *parameters* of a model. It contains a *step()* method to modify the model parameters according to the optimizer's update rule. The *SPSA* algorithm is implemented in the *SPSAOptimizer* class.
- *Dataset*: A class that provides functionality for easy management and manipulation of datasets, including batching, shuffling, and preparation based on the selected backend.

- *Trainer*: A trainer implements the *quantum machine learning* routine given a specific backend, using a *loss function* and an *optimizer*. Concrete implementations include the *QuantumTrainer* class.

The process of training a model involves the following steps:

(1) Create a *TketModel*.
(2) Create a *QuantumTrainer*, passing to it the model, a *loss function*, and a *SPSAOptimizer*.
(3) Create a *Dataset*, from PQCs, for training, and optionally, one for evaluation.
(4) Train the model by handing the dataset to the *fit()* method of the QuantumTrainer.

Once a TketModel is trained, it can be called with *unseen sentences* in the form of PQCs to make predictions.

## 5 Variational Quantum Text Classifiers

Many NLP tasks require the handling of *text consisting of multiple sentences*, which is not supported in lambeq / DisCoPy. A good example is sentiment classification of customer feedback or product review.

Text in general can have all sorts of sentence structure. Even the ordering of sentences may or may not be significant in a text. This makes the handling of text complicated, i.e., there are no grammatical rules to follow, unlike sentence. Some of the typical types of *text structure* include:

- *Description*: Shows the details of an event, person, place, or object.
- *Explanation*: Shows a set of facts and, possibly, their context, causes, and consequences.
- *Sequence-Time*: Describes how things happen or work in chronological order.
- *Problem-Solution*: Presents a problem, including why there is a problem, followed by one or more possible solutions.
- *Persuasive*: States a position and argues a reason, or set of reasons, in support of the position.
- *Cause-Effect*: Shows the connection between what has happened and its impact or result.
- *Compare-Contrast*: Shows how two or more things are alike and/or how they are different.

It can be seen that *sentence ordering* is significant for some types of text structure, e.g., sequence-time, but not for other types, e.g., description.

In the following we discuss three potential approaches to *variational quantum text classifiers (VQTCs)*, in line with VQSCs discussed previously. The first is a *weighted bag-of-sentences* approach which treats text as a group of independent sentences with task-specific sentence weighting. The second is a *coreference resolution* approach which treats text as a consolidation of its member sentences with coreferences among them resolved. Both approaches are based on the *DisCoCat* model. The third approach, on the other hand, is based on the *DisCoCirc* model which considers both ordering of sentences and interaction of words in composing *text meaning* from word and sentence meanings.

## 5.1 Weighted Bag-of-Sentences Approach

In DisCoCat, each *sentence* in a text is a *state*, i.e. it has a single output of sentence-type and no input [15]. To compose the *text meaning* from word and sentence meanings, the only thing one can do is to take the *conjunction* of all sentences.

We, therefore, treat *text* as a group of *independent sentences*. For each group (text), the member sentences have *task-specific weighting*, normalized to 1 within the group. For *text classification*, each member sentence within a group is assigned the *same topic* of the group (text). We *extend* the sentence classification algorithm used by VQSC for text classification:

- Given N groups (samples) of text with each *group j* ∈ {0, ..., N-1}, the sentences in each group is denoted $\Gamma_j$, i.e., $S \in \Gamma_j$ and $\Gamma = \{\Gamma_j\}$.
- The sentences in each group has a normalized *weight* $W_S$: $\Sigma_{S \in \Gamma_j} W_S = 1$.
- The *categorical cross-entropy loss function* is now: $C(\Omega) = \Sigma_j \Sigma_{S \in \Gamma_j} W_S L(i, S) \cdot \log P(i, C_S)$

The following is an example movie review text consisting of independent sentences:

- The actors are good.
- The directing is ok.
- The screenplay is terrible.

Assuming the movie review is rated 'poor', after training the third sentence should result in the highest weight, and the first sentence the lowest.

Not that this approach applies to text that has no *coreference* [17]. For text with coreference, *coreference resolution* should be applied first, as discussed in the next subsection.

## 5.2 Coreference Resolution Approach

*Coreference resolution* [17] is the task of finding all expressions, i.e., *coreferences*, that refer to the same *entity* in a text. This is essential for many NLP tasks based on text, e.g., machine translation. The sequence of sentences in a text forms a *discourse*. References are usually of two kinds: endophoric and exaphoric. *Endophoric reference* refers to an entity that appears in the discourse, while *exaphoric reference* refers to an entity that does not appear in the discourse. Our focus is on *endophoric reference*. Two well-known libraries that can be used for coreference resolution are: *StanfordCoreNLP* and *neuralcoref*.

The following is an example text with coreferences:

1. Alice is a dog.
2. Bob is a person.
3. She bites him.

After coreference resolution, the *consolidated sentence* looks like:

- Alice, a dog, bites Bob, a person.

Our approach is to treat text as a consolidation of its member sentences with coreferences among them resolved, i.e., all *coreferenced sentences* are merged into a single *consolidated sentence*. The text then consists of one or more consolidated sentences (some may be the original sentences if they have no coreference.) The consolidated sentences can be considered *independent sentences*. Therefore, the *weighted bag-of-sentences* approach discussed previously can be applied to them.

## 5.3 DisCoCirc Model of Text Meaning

*DisCoCirc (Circuit-shaped Compositional Distributional)* [15] defines how *sentences* interact in *text* in order to produce the meaning of that text. In DisCoCirc *word meanings* correspond to a *type*, and the states of this type can *evolve*. Sentences are *gates* within a *circuit* (for text) which update the variable meanings of those words. The compositional structure are *string diagrams* (for sentences) representing information flows, and an entire text yields a single *text circuit* in which word meanings lift to the meaning of the entire text. Both the compositional formalism and text meaning model are highly quantum-inspired.

DisCoCirc has the following notable features [16]:

- it captures *linguistic `connections'* between word and sentence meanings;

- word meanings get *updated* as text progresses;
- *sentence types* reflect the words that have their meaning updated;
- it applies to *general text*, not just sentences.

The following example text has linguistic 'connections' between words and sentences, and in which word meanings (states) get updated as the text progresses (see [15] for the composition / drawing of its *text circuit*):

1. Alice is a dog.
2. Bob is a person,
3. Alice bites Bob,

*DisCoCirc* makes fundamental modification of *DisCoCat* since a sentence in DisCoCirc updates meanings of words, whereas all meanings are static in DisCoCat. It is not clear if DisCoCirc can be implemented in lambeq / DisCoCat without breaking DisCoCat.

# 6 Conclusion

Variational quantum sentence classifiers are well supported in lambeq / DisCoPy, based on the DisCoCat model of sentence meaning. Many NLP tasks, however, require the handling of text consisting of multiple sentences, which is not supported in lambeq / DisCoPy.

We discuss three potential approaches to variational quantum text classifiers. The first is a weighted bag-of-sentences approach, and the second is a coreference resolution approach. Both approaches are based on the DisCoCat model and should be implementable in lambeq / DisCoCat. The third approach, on the other hand, is based on the DisCoCirc model which makes fundamental modification of DisCoCat. It is not clear if DisCoCirc can be implemented in lambeq / DisCoCat without breaking DisCoCat.

**Acknowledgement:** Thanks to my wife Hedy (郑期芳) for her support.

## Appendix A. Category Theory

*Category theory* [4] is the mathematical study of (abstract) algebras of *functions*. Category theory has wide-ranging applications. In fact, it turns out to be a kind of *universal mathematical language*. As a result, it tends to reveal certain connections between different fields – in our case between *natural language processing (NLP)* and *quantum computing*.

A *category* is an "algebra," consisting of *objects* A,B,C, . . . and *arrows* f : A → B, g : B → C, . . ., that are closed under *composition* and satisfy certain *conditions* typical of the composition of functions. A category consists of the following *data*:

- *Objects*: A,B,C, . . .
- *Arrows*: f, g, h, . . .
- For each arrow f there are given objects:
    dom(f), cod(f)
  called the *domain* and *codomain* of f. We write:
    *f : A → B*
  to indicate A = dom(f) and B = cod(f).
- Given arrows f : A → B and g : B → C, i.e. with:
    *cod(f) = dom(g)*
  there is given an arrow:
    *g ∘ f : A → C*
  called the *composite* of f and g.

- For each object A there is given an arrow:
  $1A : A \to A$
  called the *identity arrow* of A.

and *axioms*:

- *Associativity*: $h \circ (g \circ f) = (h \circ g) \circ f$
- *Unit*: $f \circ 1A = f = 1B \circ f$

## A.1 DisCoPy Implementation of Category Theory

*DisCoPy* [7] is a Python implementation of category theory for *linguistic* applications. In Python, the *data* of a category can be translated into Python code, but *axioms* can only be translated into test cases. The *data* for a category is given by a tuple:

$C = (C0, C1, dom, cod, id, then)$

where:

- C0 and C1 are classes of *objects* and *arrows* respectively,
- dom, cod : C1 $\to$ C0 are functions called *domain* and *codomain*,
- id : C0 $\to$ C1 is a function called *identity*,
- then : C1 × C1 $\to$ C1 is a partial function called *composition*, denoted by $\circ$. We denote the composition *then(f, g)* by $f * g$, translated to $f >> g$ or $g << f$ in Python. (Note that $f * g$ is the same as $g \circ f$.)

The *axioms* for the category C are the following:

- $id(x) : x \to x$ for all objects x $\varepsilon$ C0,
- for all arrows f, g $\varepsilon$ C1, the *composition* $f * g$ is defined
  iff cod(f) = dom(g),
  moreover we have :
  dom(f) $\to$ cod(g),
- id(dom(f)) * f = f = f * id(cod(f)) for all arrows f $\varepsilon$ C1,
- $f * (g * h) = (f * g) * h$ whenever either side is defined for f, g, h $\varepsilon$ C1.

The DisCoPy implementation of a category is a pair of classes *Ob* and *Arrow* for objects and arrows, together with four methods *dom, cod, id* and *then*. The *Functor* class is a named tuple with two attributes *ob* and *ar* for its object and arrow class respectively. *Diagram* (for string diagram) is a subclass of Arrow, and *Circuit* (for quantum circuit) is a subclass of Diagram.